\documentclass{article}

\usepackage[preprint,nonatbib]{nips_2018} %

\usepackage[utf8]{inputenc} %
\usepackage[T1]{fontenc}    %
\usepackage{hyperref}       %
\usepackage{url}            %
\usepackage{booktabs}       %
\usepackage{amsmath, amsfonts, bm}  %
\usepackage{nicefrac}       %
\usepackage{microtype}      %
\usepackage{graphicx, caption, subcaption}
\graphicspath{{images/}}
\usepackage{siunitx}
\usepackage{gensymb, textcomp}

\newcommand{\x}{\mathbf{x}}

\newcommand{\p}{\mathbf{p}}

\usepackage{acro}
\acsetup{first-long-format=\itshape\/}  %
\DeclareAcronym{MRI}{short={MRI}, long={magnetic resonance imaging}}
\DeclareAcronym{SSM}{short={SSM}, long={statistical shape model}}
\DeclareAcronym{ASM}{short={ASM}, long={active shape model}}
\DeclareAcronym{PDM}{short={PDM}, long={point distribution model}}
\DeclareAcronym{CPD}{short={CPD}, long={coherent point drift}}

\usepackage[
    backend=biber,
    style=ieee,
    sortlocale=de_DE,
    natbib=true,
    maxbibnames=9,
    doi=false,%
    giveninits=true
]{biblatex}
\title{Deep Morphing: Detecting bone structures in fluoroscopic X-ray images with prior knowledge}

\author{
Aaron~Pries\qquad Peter~J.~Schreier\qquad Artur~Lamm\qquad Stefan~Pede\\
Signal \& System Theory Group\\
Paderborn University, Germany\\
\AND%
Jürgen~Schmidt\\
Klinik für Unfallchirurgie, Orthopädie, Plastische und Handchirurgie\\
Klinikum Augsburg, Germany
}
\begin{document}

\maketitle

\begin{abstract}
	We propose approaches based on deep learning to localize objects in images when only a small training dataset is available and the images have low quality. That applies to many problems in medical image processing, and in particular to the analysis of fluoroscopic (low-dose) X-ray images, where the images have low contrast. We solve the problem by incorporating high-level information about the objects, which could be a simple geometrical model, like a circular outline, or a more complex statistical model. A simple geometrical representation can sufficiently describe some objects and only requires minimal labeling. Statistical shape models can be used to represent more complex objects. We propose computationally efficient two-stage approaches, which we call \emph{deep morphing}, for both representations by fitting the representation to the output of a deep segmentation network.
\end{abstract}

\section{Introduction}

In medical image processing, we often deal with problems where the images are challenging to analyze due to their low quality and only a limited training dataset is available. In these cases it can be helpful to incorporate prior knowledge. Examples are localization problems where the objects can be approximated by a simple geometrical shape or a statistical model. That is the case for X-ray images of the femur (thigh bone), which is the main focus of this paper. Localizing the femur in an X-ray image is helpful for many medical applications~\cite{Tian2003,Tannast2007,Wang2011,Sutter2012,Bojan2013,Regling2014,Lindner2015} to identify pathologies or assist during surgery. To do that, an algorithm needs to identify the anatomical parts of the bone. In the case of the femur that could be the precise location of the femoral head, the greater trochanter, etc.

\begin{figure}[tb]
	\centering
	\begin{subfigure}[t]{0.48\textwidth}
		\centering\includegraphics[width=\textwidth]{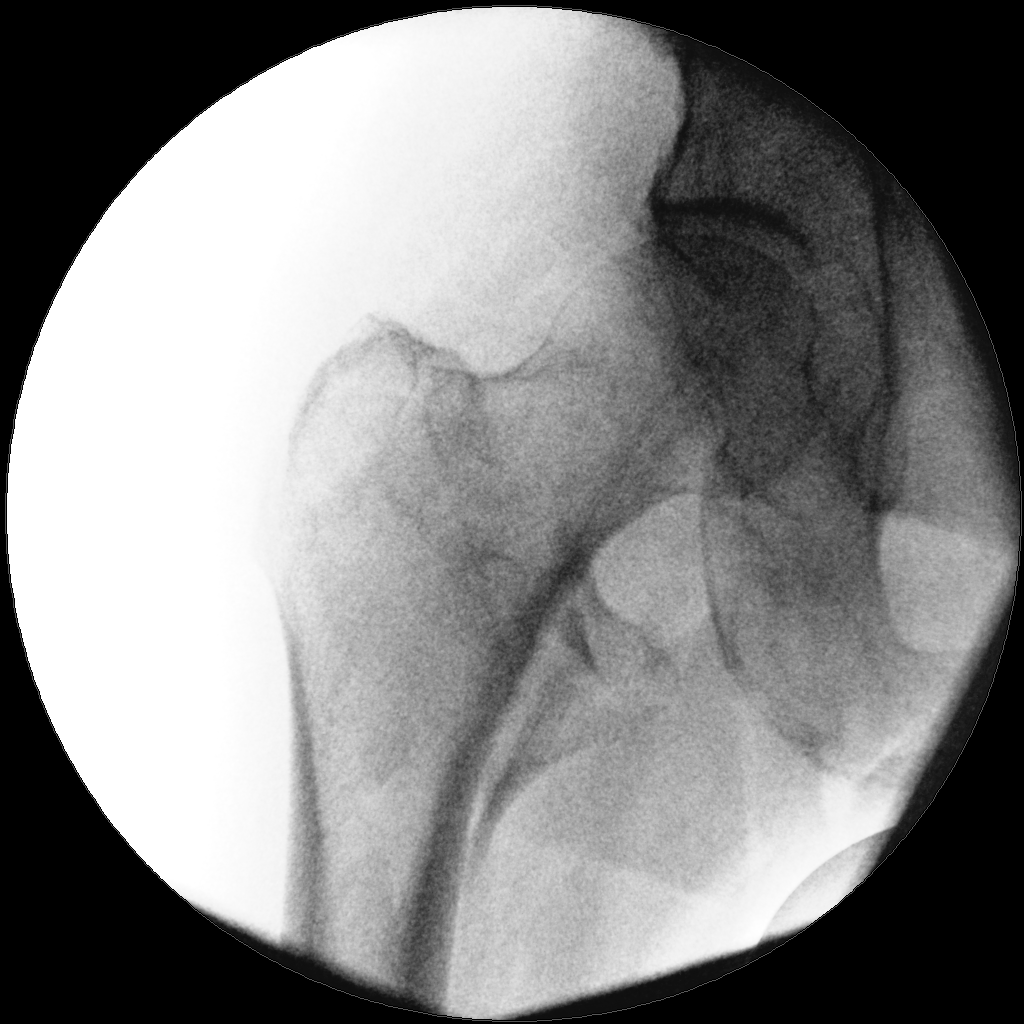}
	\end{subfigure}%
	\quad
	\begin{subfigure}[t]{0.48\textwidth}
		\centering\includegraphics[width=\textwidth]{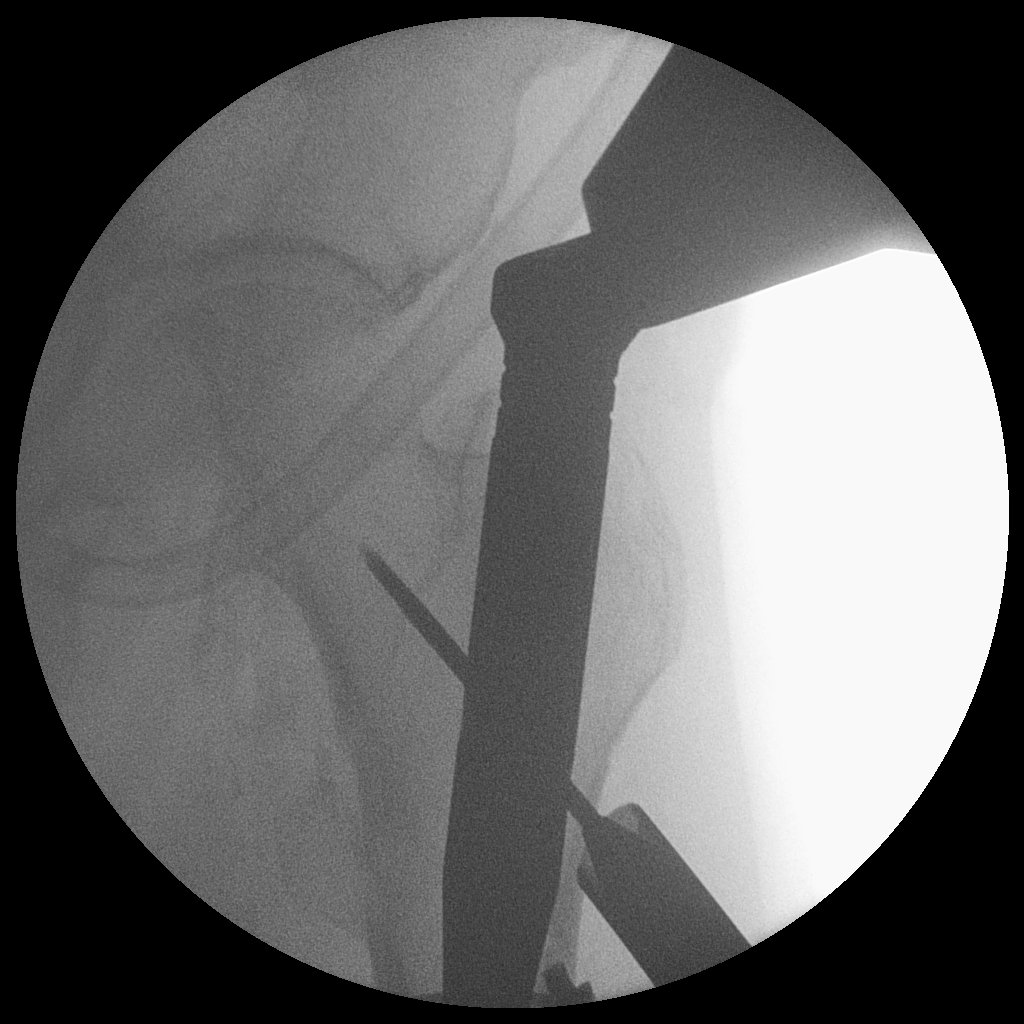}
	\end{subfigure}
	\caption{Fluoroscopic X-ray images with different contrast showing the femur. The black border is due to the disk-shaped detector of the C-arm. The right image also shows an implant.}%
	\label{fig:example_image}
\end{figure}

X-rays taken intraoperatively are often fluoroscopic X-rays in order to limit the radiation exposure of surgeons and operating room staff. Fluoroscopic X-rays are taken with a comparatively low radiation dose, which results in low SNR and contrast (see~\autoref{fig:example_image}), making their automatic processing challenging. The scans are often made with so-called C-arms, which can be easily repositioned and rotated during the surgery. However, this also means that the images do not have a standardized appearance. In particular, scale, image-plane rotation, viewing angle, and contrast/brightness vary in a wide range and increase the search space for the object detection. During surgery, the femur is furthermore often occluded by implants and surgical tools, as shown in \autoref{fig:example_image}.%

Algorithms for the detection of the femur were already proposed for different 3D modalities such as MRI (\cite{Deniz2017}), CT (\cite{Krcah2011}), and 2D radiographic X-rays (\cite{Behiels2002,Chen2005,Gamage2010,Lindner2013,Xie2014}). Detectors for parts of the femur in fluoroscopic X-rays were presented in~\cite{Dong2009,Gamage2010,Wang2011,Eguizabal2017}. In this paper, we develop a new approach that uses deep learning combined with prior information about the shape of the bone. We will use fluoroscopic X-ray images for our evaluation, but our approach can be applied to other 2D and 3D modalities as well.

A classical solution to localize anatomical points on a contour is the \ac{ASM} algorithm~\cite{Cootes1992a}. For the \ac{ASM} algorithm, the variation of the outline and the gray-level appearances near the outline are learned from training data. However, the standard \ac{ASM} only employs a low-capacity model and fluoroscopic X-ray images have a low SNR, which means that simple models of the appearance are not sufficient for an accurate localization. Similar to the techniques in~\cite{Cheng2016, Cheng2017,Zadeh2017} a more complex patch-based neural network can be used for a refinement step. However, such an approach is inherently slow, and in~\cite{Cheng2017}, the authors report a computation time of 8 minutes.%

The state of the art for localizing the femur in \emph{radiographic} (i.e., higher-dose and thus higher-quality) X-rays uses random forests, which can achieve high capacity, and combines them with constrained local models~\cite{Lindner2013}. However, the random forests are used in a part of the algorithm where they have a strong impact on runtime.

An end-to-end solution to predict the coordinates of anatomical points with a neural network usually requires many training images and thus does not work for many medical imaging applications. Pixel-wise segmentation networks, on the other hand, have been shown to work well even with limited amounts of training data~\cite{Ronneberger2015,Long2015}. They can be used to find the outline of objects, but they do not label particular points on the outline. Instead, we propose to fit the geometrical or statistical representation of the model to the output of a segmentation network, a process we call \emph{deep morphing}. This allows us to localize the objects as well as labeling specific points on their outline.

\begin{figure}[tb]
	\centering
	\begin{subfigure}[t]{0.65\textwidth}
		\centering\includegraphics[height=4cm]{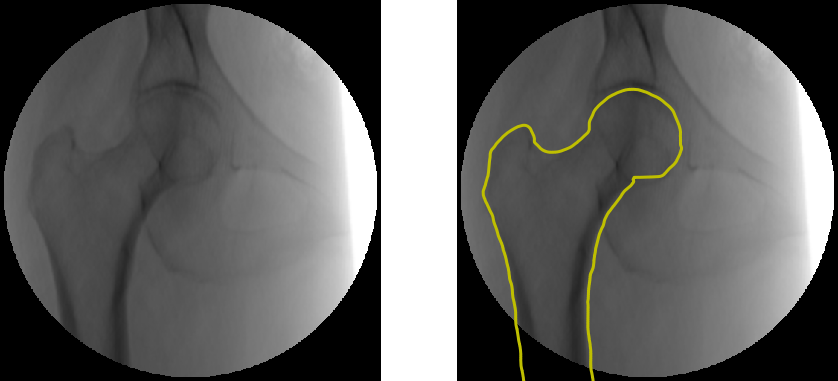}
	\end{subfigure}%
	\quad
	\begin{subfigure}[t]{0.32\textwidth}
		\centering\includegraphics[height=4cm]{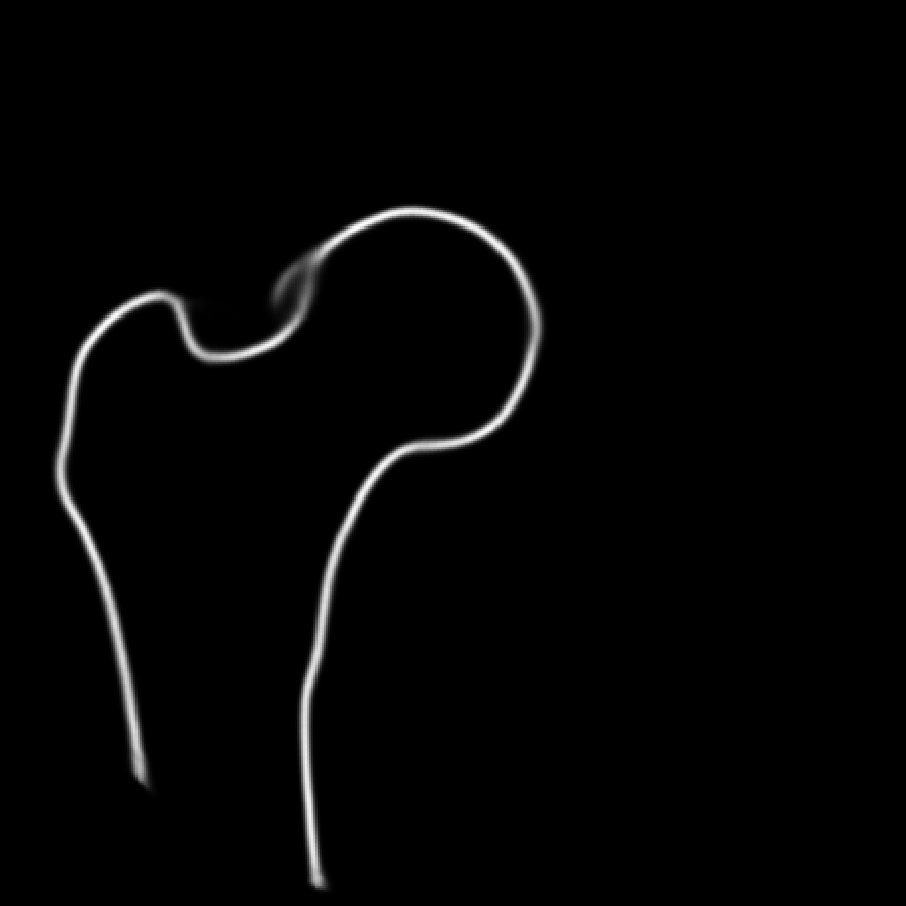}
	\end{subfigure}
	\caption{Simulated C-arm image of the proximal part of the femur (left), the true outline overlaid in yellow (middle), and the prediction of a trained segmentation network (right).}%
	\label{fig:simulated_with_gt}
\end{figure}

Segmentation networks perform pixel-wise classification, typically by means of a series of convolution layers, downsampling, and upsampling operations. In the case of a single (foreground) class, the last operation in the network is a sigmoid activation function. The activation value is a measure of confidence for the presence of the foreground class, as shown in \autoref{fig:simulated_with_gt},
which can be thresholded for a hard assignment.

The outline of this paper is as follows. In \autoref{sec:head} we introduce the first problem, where we localize the femoral head, and in \autoref{sec:femur} we discuss the problem of localizing keypoints on the entire femur. In \autoref{sec:evaluation} we evaluate the performance of our proposed solutions on a dataset of fluoroscopic X-ray images. Since the number of images in our dataset is relatively small, we train our networks with simulated images, which is explained in the appendix. %
\section{Detecting the femoral head}%
\label{sec:head}
In this section, we focus on the detection of the femoral head, which is the part of the femur that meets the pelvis. This specific part of the femur is frequently of interest in medical applications~\cite{Tannast2007,Wang2011,Sutter2012,Bojan2013,Regling2014}. The femoral head is roughly spherical~\cite{Hammond1967,Petersik2011}, and in 2D X-ray images its contour is frequently approximated by a circle~\cite{Sutter2012, Tannast2007}. Thus we are interested in the center coordinates $\mu_x$ and $\mu_y$, and the radius $r$. %

A straightforward solution to this problem using a DNN is a regression network that estimates the circle parameters \( \bm\theta = {[\mu_x, \mu_y, r]}^T \) and is trained to minimize the cost function
\begin{equation}
	\|\bm\theta - \hat{\bm\theta} \|^2.
	\label{mse}
\end{equation}
This is similar to the localization through bounding-box regression used, for example, in~\cite{Simonyan2014}. A second solution to this problem is an R-CNN approach~\cite{Ren2015a}, where one network proposes regions of interest and another network classifies them and predicts a bounding box. Such a solution was employed in~\cite{He2015} for a localization problem with many classes. In our case, we know that exactly one instance of a femoral head will be visible. Hence we are interested in a different approach for the localization of the femoral head.

We propose the following two-stage approach: First, a segmentation network predicts the outline of the femoral head. Given the 2D output $O(x,y)$ of the the first stage at (pixel) locations $x$ and $y$, we then identify the set of foreground pixels
\begin{equation}
	P_\mathrm{fg} = \{(x, y) \mid O(x, y) > 0.5 \}
	\label{P_fg}
\end{equation}
and solve the following nonlinear least-squares fit of the geometric distances~\cite{Gander1994}:
\begin{equation}
	\min_{\mu_x, \mu_y, r} \sum _{(x,y) \in P_\mathrm{fg} } {\left(\sqrt{ {(\mu_x - x)}^2 + {(\mu_y - y)}^2} - r \right)}^2.
	\label{geometric_fit}
\end{equation}
To avoid an iterative solution, we can also parameterize the circle as $x^2 + y^2 + Bx + Cy + D$ and minimize the \emph{algebraic} distance~\cite{Chernov2005}:
\begin{equation}
	\min_{B, C, D} \sum _{(x,y) \in P_\mathrm{fg} } {\left(x^2 + y^2 + Bx + Cy + D\right)}^2.
	\label{algebraic_fit}
\end{equation}
For this problem, a closed-form solution exists~\cite{Chernov2005}.

This approach can easily be generalized to other contours in 2D or surfaces in 3D by replacing the cost function of the second stage to reflect the object of interest. This could be useful for example %
for glaucoma screening, where the optic disc in fundus images can be approximated by an ellipse~\cite{Fu2018}. Our approach can also be used to localize the femoral head in 3D CT scans, by performing a 3D segmentation followed by a fit of a sphere. %
\section{Detecting the entire proximal femur}%
\label{sec:femur}
In this section, our goal is to identify the contour of the femur and to label its anatomical parts. Algorithms for such a task can be classified into general \emph{keypoint localization} algorithms, where the spatial relationship between keypoints is often modeled implicitly, and approaches that use \acp{SSM}. Many variations for both types of algorithms have been presented for problems like face recognition, human pose estimation, and the detection of bodily organs in medical imaging. %
Here, we are interested in a novel combination of \acp{SSM} and deep learning that allows a fast and accurate detection of the proximal femur.

A simple instance of an \ac{SSM} is a model that captures the distribution of the location of specific points on the shape, i.e\ a \ac{PDM}. In its typical formulation (\cite{Cootes1992}), a $2N$-dimensional vector is constructed with the coordinates of the $N$ points: \( \x = {[x_1, y_1,\ \dots, x_N, y_N]}^T \). These points are then approximated by the linear model
\begin{align}
	\x  & \approx \bar{\x} + \sum_{i=1}^{M} \p_i b_i,\quad \mathrm{with} \\
	b_i & = \p_i^T(\x - \bar{\x}),
\end{align}
where $\bar{\x}$ is the sample mean of $\x$ and the $\p_i$'s are the orthonormal eigenvectors of the sample covariance matrix of $\x$, sorted by decreasing eigenvalues $\lambda_i$. The $\p_i$'s can be interpreted as the modes of the shape, and the $b_i$'s measure the strength of the modes. For $M=2N$, the approximation becomes exact, but in practice not all modes are used to avoid overfitting the training dataset. Choosing $M$ such that a sufficiently large fraction of the variance is retained works well in practice~\cite{Cootes1995}. Alternatively, $M$ can be chosen automatically based on information-theoretic criteria~\cite{Davies2002,Eguizabal2018}.

This \ac{PDM} also allows for regularization of shapes which are \enquote{far} away from the distribution observed in the training phase. In such a case some $b_i$'s are typically large in magnitude, and the reconstruction can be regularized, for example by clipping them to \( \pm 3 \sqrt \lambda_i\)~\cite{Cootes1992}.

A \ac{PDM} is typically fit to a shape observed in an image with the \ac{ASM}~\cite{Cootes1992a} algorithm or one of its variants. Given an initial position of the shape, these algorithms iteratively fit the shape model to the image by first moving the current points to points on the desired outline based on the information in the image. Then these points are constrained to conform to the typical variations observed in the training data of the \ac{PDM}. These two steps are repeated until convergence.

In preliminary tests, we experimented with a two-class segmentation network that predicts whether or not a pixel is a boundary pixel of the femur. We observed that such a network is indeed able to identify the entire contour of the femur in low-dose X-ray images. Hence we propose to use a segmentation network as a first stage. Next, we initialize and fit an \ac{PDM} to the \emph{output map} of the network. We will now explain these steps in more detail.

\subsection{Initialization of the \ac{PDM}}
The \ac{ASM} algorithm relies on an initial placement of the mean shape in the image. Without prior information, this is quite difficult in fluoroscopic X-ray images since they can be arbitrarily rotated and have a low SNR and contrast.\ If a segmentation network is able to identify the contour as shown in \autoref{fig:simulated_with_gt}, this can be simplified.

As a first step, the coordinates of the mean shape of the \ac{PDM} need to be scaled, rotated, and shifted such that they are close to the shape observed in the image. The segmentation network yields a probability map for the outline of the bone. However, it does not indicate which points on the outline correspond to particular points of the \ac{PDM}. Hence we need to simultaneously find the corresponding points on the observed shape and a rigid transformation.

This can be achieved with the \ac{CPD} algorithm~\cite{Myronenko2010}. \ac{CPD} finds a transformation $\mathcal{T}$ of the points of the mean shape $(x_n, y_n)$, and estimates the posterior probability $p(n \mid \tilde{x}_m, \tilde{y}_m )$ of the $n$th transformed point of the mean shape given the observed point $(\tilde{x}_m, \tilde{y}_m) \in P_\mathrm{fg}$, as defined in~\eqref{P_fg}.

\begin{figure}[tb]
	\centering
	\begin{subfigure}[t]{0.48\textwidth}
		\centering\includegraphics[height=5cm]{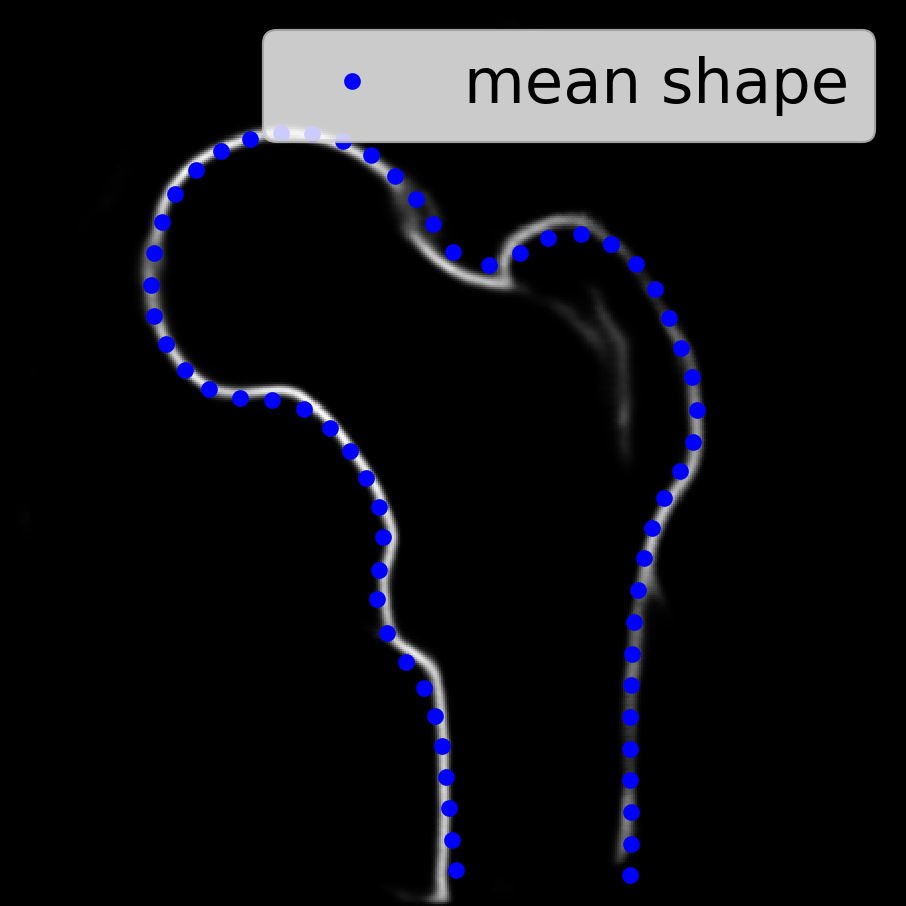}
		\caption{Initialization of the model: Prediction of a DNN with the mean shape fitted to the foreground pixels.}%
		\label{fig:prediction_with_fit}
	\end{subfigure}%
	\quad
	\begin{subfigure}[t]{0.48\textwidth}
		\centering\includegraphics[height=5cm]{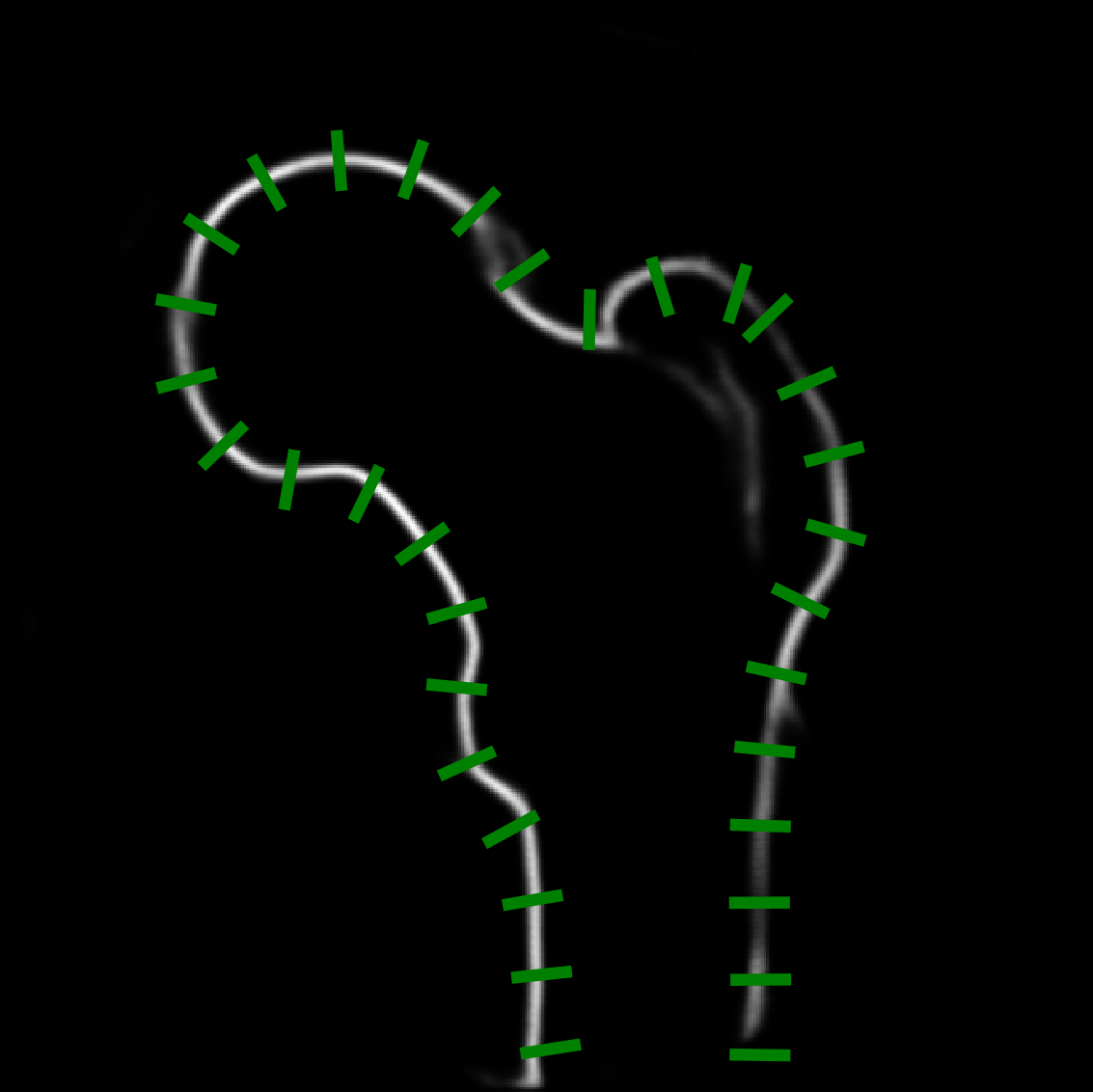}
		\caption{Update along lines orthogonal to the current shape.}%
		\label{fig:update}
	\end{subfigure}
	\caption{Steps during the shape model fit.}
\end{figure}

To initialize the \ac{PDM} with \(\x_\mathrm{init} = [\bar{x}_1, \bar{y}_1, \dots, \bar{x}_N, \bar{y}_N]^T\), we thus use the transformed points of the mean shape:
\begin{equation}
	(\bar{x}_n, \bar{y}_n) = \mathcal{T}(x_n, y_n).
\end{equation}
An example can be seen in \autoref{fig:prediction_with_fit}. These initial points will not match the points from the segmentation network.
This initial shape is then fit to the contour of the segmentation network such that we reach a good tradeoff between fitting the observed data and staying close to shape variations observed in the training data. This will be described in \autoref{sec:fit}.

One final comment on the robustness of this initialization is in order: The \ac{CPD} algorithm breaks down if one point cloud is rotated too much~\cite{Myronenko2010}. This can be dealt with by performing the registration with differently rotated versions and choosing the best match. The same applies if one of the point clouds is reflected.

\subsection{Fitting the shape model}%
\label{sec:fit}
In the \ac{ASM} algorithm~\cite{Cootes1992a}, the shape model is updated by finding better points on lines orthogonal to the current shape. This is done by comparing the observed gray-level profile on the orthogonal line with those observed in the training data. But since our images have such low SNR, local gray-level information is not very discriminative. Hence we do not update the shape model based on the original image but instead fit it to the output of the segmentation network. More precisely, we use the two-dimensional output of the last sigmoid-layer, i.e.\ \emph{before} applying thresholding for classification. This output measures the confidence of the network and thus contains more information that can be exploited for fitting.

\begin{figure}
	\centering
	\includegraphics[width=0.5\textwidth]{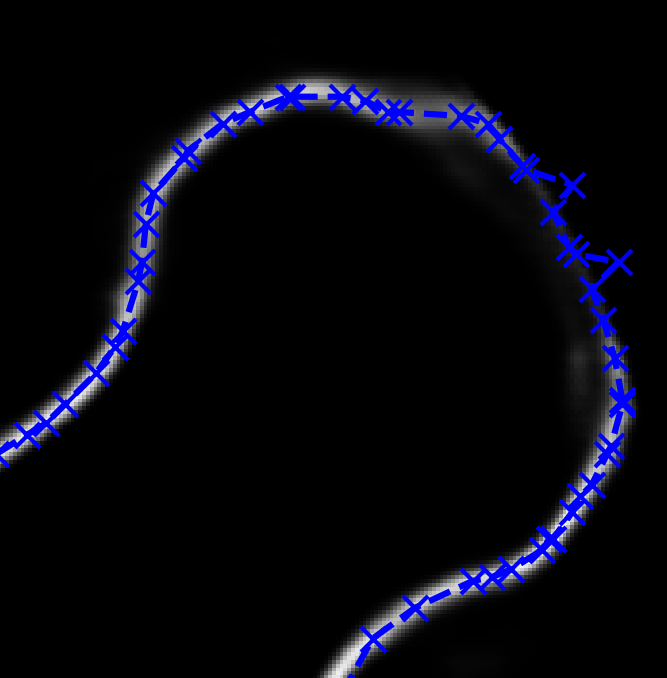}
	\caption{Irregular shape in a region of low confidence. Such a shape can be regularized with the \ac{PDM} constraint.}%
	\label{fig:constraint}
\end{figure}

Given a segmentation with an initial placement of the mean shape, we update the points of the shape with the position of the maximum confidence along the orthogonal lines of the shape (\autoref{fig:update}). This is followed by the constraint of the shape (\autoref{fig:constraint}). These steps are then repeated for a few iterations.

\section{Evaluation}%
\label{sec:evaluation}
We have an in-house dataset of 211 fluoroscopic X-rays from Klinikum Augsburg, showing the proximal part of the femur. These X-rays are taken intraoperatively during surgeries treating hip fractures, hence many of them contain surgical tools and implants. The number of images in our database is insufficient to train neural networks, thus we will only evaluate our algorithms on this data and use simulated data for training (\autoref{sec:simulator}).

For the network architecture, we choose the FusionNet~\cite{Quan2016}, which is a variant of the U-Net~\cite{Ronneberger2015} but with residual connections. %
During training we rely on augmentation to control overfitting: We use random rotations of up to 360°, reflections, additive spatially colored noise, and gamma transformations. We furthermore block square regions in the training images, which is a simple model for the occlusions by implants and tools in the real X-ray images in our database. The networks are trained with a cross-entropy loss for 80 epochs, and the learning rate is decayed by a factor of $0.1$ after 60 epochs.

We are interested both in detecting only the femoral head (which for some applications is all that is needed) and in detecting the entire outline of the femur. We now evaluate the performance of our deep-morphing approaches presented in Sections~\ref{sec:head} and~\ref{sec:femur}.
\subsection{Femoral head detection}
We first consider the detection of the femoral head (see \autoref{sec:head}). Since it is located inside the pelvis, the femoral head is particularly difficult to identify in fluoroscopic X-ray images. We annotated the femoral heads in our dataset, and the average time of $\SI{54}{\second}$ per circle-annotation shows how difficult this problem is.

We trained a segmentation network to predict the outline of the femoral head region in simulated images. Because we derive the ground truth from the simulated data, the target outline does not resemble a perfect arc. In other scenarios, the target map could consist of the points on a circle or an arc, which requires less manual annotation compared to a pixel-wise outline.

\begin{figure}[tb]
	\centering
	\begin{subfigure}[t]{0.48\textwidth}
		\centering\includegraphics[width=\textwidth]{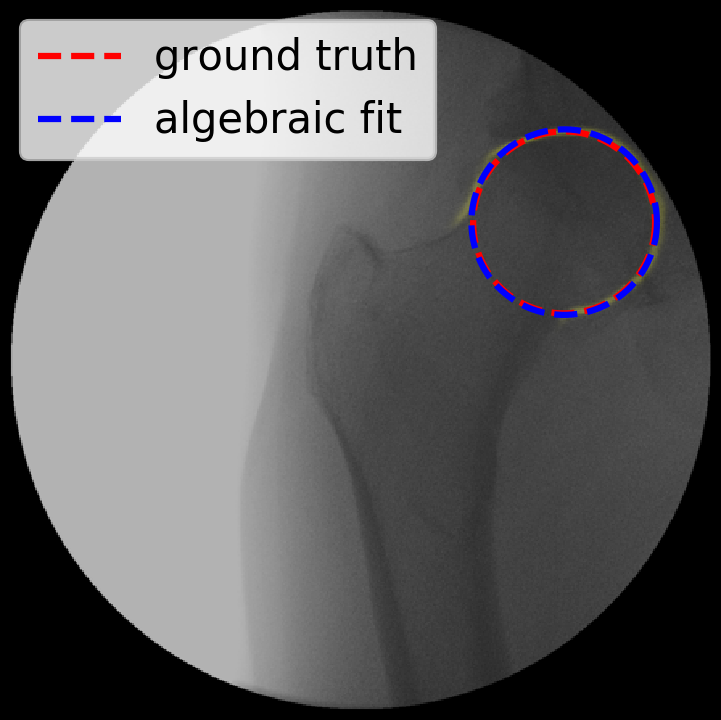}
		\caption{Estimated circle is nearly identical to the ground truth.}%
		\label{fig:correct_circle}
	\end{subfigure}%
	\quad
	\begin{subfigure}[t]{0.48\textwidth}
		\centering\includegraphics[width=\textwidth]{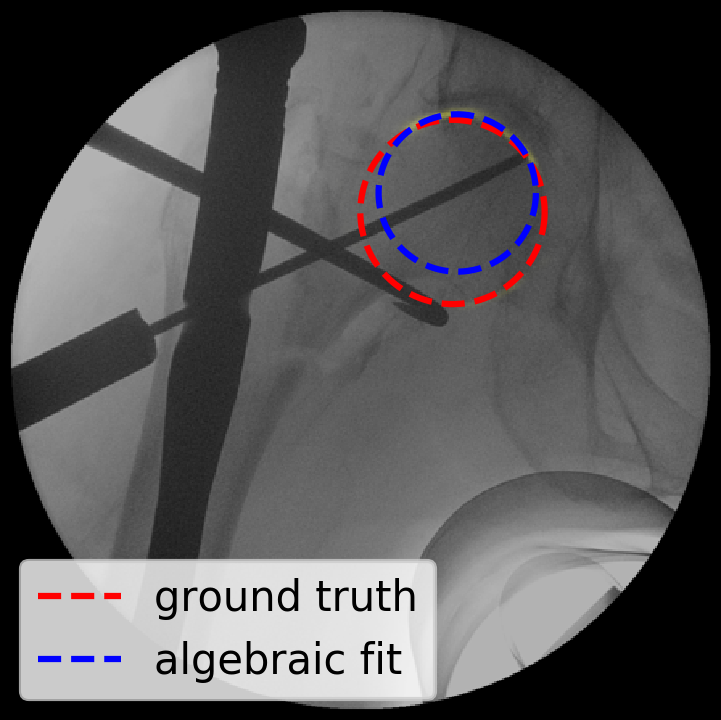}
		\caption{Tools and implants are visible and lead to an inaccurate circle.}%
		\label{fig:incorrect_tools}
	\end{subfigure}
	\caption{Examples of femoral head detections.}
\end{figure}

For evaluation, we predict the outline with the trained network and fit the circle to the outline as proposed in \autoref{sec:head}. Given a bigger dataset of real X-ray images, this procedure could also include fine-tuning the network to improve generalization. In our case, the two-stage approach generalizes quite well on real images, especially when there are no occlusions by tools or implants. In that case, the RMSE, defined as \(\|\bm\theta - \hat{\bm\theta} \|\), is 3.8 pixels on average for both the geometric~\eqref{geometric_fit} and the algebraic fit~\eqref{algebraic_fit} when working with images of size 448 $\times$ 448.\footnote{For our problem, the algebraic and geometric fits usually yield very similar results, so we choose the algebraic fit because it is faster.} An example of such a segmentation is shown in \autoref{fig:correct_circle}.

If there are implants visible in the X-ray image, the segmentation network tends to predict fewer foreground pixels. As a result, the fitted circles become less accurate, as shown for example in \autoref{fig:incorrect_tools}. In some cases, the network predicts very few circle pixels and the fitted circles become very inaccurate. However, we can avoid many incorrect detections by requiring a certain number (for example 100) of foreground pixels to be predicted. Then the average RMSE on images with occlusions by tools is 8.5 pixels, but fine-tuning on real images with occlusions should help close this gap.

\subsection{Proximal femur detection}
\begin{figure}[tb]
	\centering
	\begin{subfigure}[t]{0.48\textwidth}
		\centering\includegraphics[width=\textwidth]{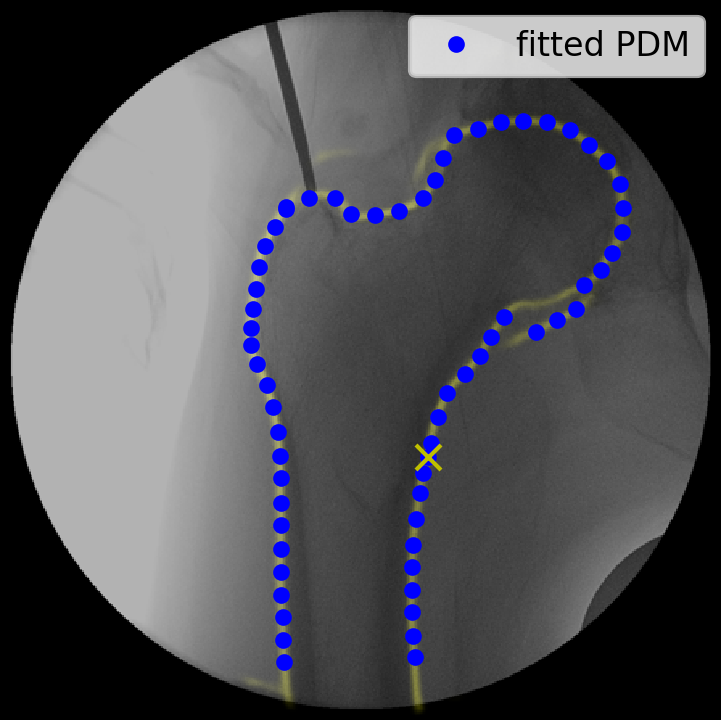}
		\caption{}%
		\label{fig:fit_real_data}
	\end{subfigure}%
	\quad
	\begin{subfigure}[t]{0.48\textwidth}
		\centering\includegraphics[width=\textwidth]{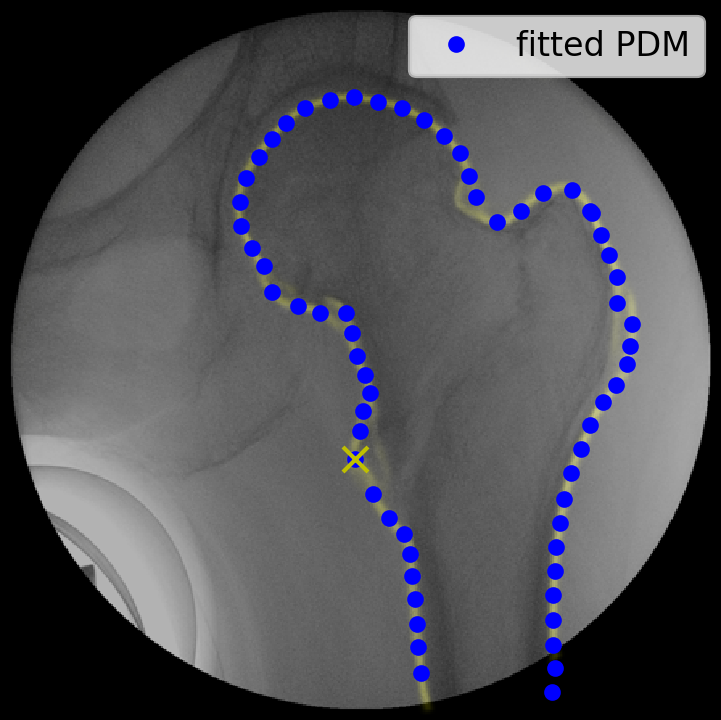}
		\caption{}%
		\label{fig:fit_real_data2}
	\end{subfigure}
	\caption{Final results of fitting a shape model to the output of a segmentation network. Our model also identifies the correspondences to anatomical parts. An example is the yellow $\times$, which represents the lesser trochanter.}
\end{figure}

We now consider the detection of the entire proximal femur (see~\autoref{sec:femur}). We do not have a dataset of real X-ray images with manually labeled \emph{landmarks}, hence we derived a 2D \ac{PDM} from a 3D surface model of the bone. An example of a real image with a fitted shape model is shown in \autoref{fig:fit_real_data}. Overall, the outline is found very accurately, except for the femoral neck region where the network prediction is ambiguous. The PDM ends up too far outside of the bone in that region, but the overall point-to-curve RMSE for a manual segmentation is still quite low with a value of $2.3$ pixels.\footnote{Since a manual annotation of points outside of the circular beam cone is not well defined, we only compute the RMSE for points inside.} In \autoref{fig:fit_real_data2}, the average point-to-curve RMSE between the predicted points and a manual segmentation of the outline is $1.6$ pixels. Here, most parts are found very accurately.

In images without occlusions by tools or implants (which we have not simulated yet in our training data), the network generalizes very well and is capable of identifying the outline. Since our 2D \ac{PDM} is a model that includes a large part of the shaft, fitting can fail when the image only shows the part of the femur without the shaft. For problems where a large model is not of interest, a shorter \ac{PDM} could be used instead.

In images with occlusions, the network does not generalize as well as in the case of detecting the femoral head because the implants typically occlude large portions of the greater trochanter and the bone marrow.
\section{Conclusion}
We have presented two-stage approaches called deep morphing to localize objects with geometrical or statistical prior knowledge. In the case of geometrical information, the model itself is very simple, which means that the manual labeling effort is quite low. Still, a simple geometrical model of a circular outline is expressive enough to solve the problem of localizing the femoral head in fluoroscopic X-ray images. If more expressive power is required, the object can be described by a statistical model, which can be fit directly to the outline predicted by a segmentation network.

As in~\cite{Unberath2018}, we also demonstrated the feasibility of training the networks on a dataset of simulated X-ray images. This allows us to circumvent the problem of limited datasets in medical image processing. To some extent, the networks generalize to real X-ray images, although effects that were not simulated (e.g.\ surgical tools) are problematic. Apart from including tools and implants in the simulated X-rays, this could also be alleviated by either fine-tuning the networks on real images or by improving the realism of the simulated images with a generative adversarial network~\cite{Shrivastava2017}.

Our deep morphing algorithm can be modified to process X-ray images acquired from different views (e.g.\ anterior-posterior and medial-lateral) or X-ray images showing different parts of a bone (e.g.\ the proximal and the distal part of the femur). This can be achieved by separate models that are morphed to the segmentation and the best fit is chosen afterwards~\cite{Cootes2000a}. Moreover, classification networks can be trained to distinguish between different views and to classify which part of a bone is visible in the image. Based on the decision of the classification network, the appropriate model is then chosen before applying deep morphing. 
\subsubsection*{Acknowledgments}
To simulate our X-ray images, we used CT scans from the SMIR dataset~\cite{Kistler2013}. The fluoroscopic X-ray dataset was provided by the Klinikum Augsburg.

\appendix
\section{Simulated fluoroscopic X-ray images}%
\label{sec:simulator}
The low number of X-ray images in our dataset is problematic when training neural networks. However, we may simulate X-ray images with the information stored in 3D CT scans. Both modalities are closely related, since they are both based on the physical principle of X-rays passing through the body. Since CT scans compute a 3D representation, this allows us to simulate the process of creating fluoroscopic X-ray images from different perspectives.

\subsection{Rendering CT scans}
Most commonly, the voxels of a CT scan are encoded with 12 bits. They may be stored with values between $0$ and $4095$, or shifted by multiples of $1024$. The exact value that was used for a particular CT scan can be found in the corresponding DICOM files. So-called CT numbers are defined in the range $[-1024 \text{~HU},3071 \text{~HU}]$, where HU are Hounsfield units. Thus, the values of the CT scan are either increased or decreased, to have all CT scans in the same range. These CT numbers can be converted to attenuation coefficients by using the formula~\citep{Vidal2016}
\begin{equation}
	\mu = \frac{CT + 1024 \text{~HU}}{1024 \text{~HU}} \cdot \mu_w~,
\end{equation}
where $\mu_w$ is the attenuation coefficient of water. Air has an attenuation coefficient of $0$, which is the lowest possible value. The highest coefficient for a 12-bit dataset is $4 \mu_w$.\\
To obtain a simulated image from a CT scan, it is necessary to define the position of the virtual focal point as well as the coordinates of the image plane, which represents the receiver of the C-arm. Depending on the desired view, they are defined such that a direct line between the focal point and the image plane would go through the 3D CT data.\\
The method we used to create a simulated X-ray image is called ``ray casting.'' As the name suggests, we cast rays from the focal point to each pixel of the image plane. Since the voxels have a spatial resolution of approximately $1$~mm, it is rather improbable to hit a voxel with a ray. However, we want to collect all the information of the CT data along the ray path. Therefore, it is necessary to interpolate the 3D dataset. We decided to use trilinear interpolation for this step because it is more accurate than the nearest-neighborhood method and faster than tricubic interpolation. For each pixel we need to sample the line between the focal point and the image plane. Since according to the design of the C-arm this distance is approximately $1000$~mm, we chose a sampling size of $2000$ in order to get all the information along the path. The total attenuation for a pixel is given by the sum
\begin{equation}
	A(x,y) = \sum_{i} \mu_i \Delta x_i~,
\end{equation}
where $\mu_i$ is the attenuation coefficient of the $i$th interpolated voxel along the path and $\Delta x_i$ is the spatial step size in mm. The last step is to convert the attenuation to an 8-bit grayscale image. A high attenuation will lead to a small intensity value, and vice versa. To cover the effect of over-exposure and to increase the contrast, we chose $2.5$~\% of all pixels to reach the highest possible intensity value. For this reason, we threshold the attenuation values accordingly. Neglecting a multiplying factor, which is linearly related to an initial intensity value, the image intensity given the attenuation is defined as~\cite{Vidal2016}
\begin{equation}
	I(x,y) = \exp (-A(x,y))~.
\end{equation}
Finally these intensities are linearly scaled to the desired grayscale range, e.g., $[20, 255]$. Using a value greater than $0$ for the lower number generally leads to more realistic images. An example can be seen in \autoref{fig:simulated_with_gt}.

\subsection{Creating ground truth data}%
\label{sec:sim_gt}
Neural networks require a ground truth labeling for the training step. In our case, we want to obtain the outer contour of the bone seen in the simulated image. To avoid segmenting the bone in all images, we segmented the femurs in 3D using the CT scans. This results in triangle mesh models where the 3D femur is defined by vertices and faces. Since we define the focal point and the image plane, it is possible to project every vertex of the mesh model onto the image plane. This leads to a dense point cloud and thus, in the next step, we create the outer contour using MATLAB's boundary function. 
\renewcommand*{\bibfont}{\small}
\printbibliography%
\end{document}